%% file: main.tex
\documentclass{article}
\usepackage[T1]{fontenc}
\usepackage{xcolor}
\usepackage[preprint]{corl_2026}
\hypersetup{hypertexnames=false,
 pdftitle={Do Agentic Policies Need to Look at Every Step?
Geometric Abstraction for Reasoning and Action},
pdfauthor={Yihang Liu, Yiming Chen, Mingzhe Zheng}}
\usepackage{amsmath,amssymb,booktabs,graphicx,tabularx,array}
\usepackage{algorithm,algpseudocode}
\usepackage{microtype}
\usepackage{stmaryrd}

\title{Do Agentic Policies Need to Look at Every Step?\\
Geometric Abstraction for Reasoning and Action}
\author{Yihang Liu$^{1}$ \quad Yiming Chen$^{2}$ \quad  Mingzhe Zheng$^{3}$\\
$^{1}$Independent Researcher\\
$^{2}$Georgia Institute of Technology\\
$^{3}$The Hong Kong University of Science and Technology\\
\texttt{yihang.liu.pc@gmail.com}\quad \texttt{ychen3868@gatech.edu}\\
\texttt{mzhengar@connect.ust.hk}}
\begin{document}
\maketitle
\begin{abstract}
Recent multimodal language models can reason directly about robot scenes, but it is
unclear whether they need to repeatedly reinterpret raw visual observations at every
step of an interaction. We propose shifting routine spatial reasoning from images to
text and symbols through a 3D geometric abstraction that is refreshed locally from
observations and read directly by the language models. Specifically, our interface
represents scene geometry via typed, evidence-aware symbolic references to object
parts, grounds them from calibrated RGB-D, and returns compact textual checkpoints
with metric residuals after each action. Symbolic task structure can thus be reused
across decisions while metric values are rebound to the current scene, with direct
visual reasoning invoked only on demand. Across offline policy synthesis and
continuous online execution, reasoning over grounded text and symbols improves
multi-step manipulation completion while substantially reducing repeated multimodal
queries, positioning explicit geometric state as an effective substrate for agentic
robot policies.
\par\vspace{2pt}\noindent\textbf{Project page:} \url{https://geometric-abstraction.github.io}
\end{abstract}
\keywords{Agentic Robotics, 3D Abstractions for Robotics, Spatial Reasoning}

\section{Introduction}
\label{sec:intro}

Recent frontier multimodal models have demonstrated increasingly strong 3D spatial reasoning and, in some cases, can directly drive robot manipulation without task-specific policy training~\citep{zhang2026unexpected,jia2026agentaspolicy,tang2025incentivizing,song2025robospatial}.
However, deploying such models in closed-loop manipulation often requires repeatedly reinterpreting visual observations as the scene evolves, even though many manipulation decisions are governed by a compact set of task-relevant geometric quantities~\citep{nasiriany2025rtaffordance,zhang2025atk}.
Repeatedly recovering these quantities from raw images introduces both reasoning uncertainty and additional computational cost: metric and structural relations remain challenging for vision-language models, whereas explicitly representing spatial information in text can improve reasoning reliability~\citep{wang2024picture,pothiraj2025capture,song2025siri,yang2025thinking}.
Moreover, visual tokens often carry redundancy, incurring unnecessary overhead during reasoning~\citep{chen2024image}.
This raises a central question: \emph{does a robot agent need to reinterpret the visual scene at every decision step?}

Existing interfaces trade off richness and compactness. Raw images retain detailed scene information but require repeated visual interpretation, while conventional perception APIs often expose only coarse object-level quantities such as centers or bounding boxes. ReKep~\citep{huang2024rekep} shows that explicit 3D keypoints and relational constraints can provide an effective interface for manipulation. Meanwhile, language-model-based robotics is moving toward persistent and structured policies: Code as Policies~\citep{liang2022code} synthesizes executable programs, ASPIRE~\citep{lu2026aspire} extends this paradigm through execution--diagnosis--repair and skill reuse, and Graph-as-Policy~\citep{chen2026gap} represents policies as modular graphs. These works make the policy itself structured and reusable, raising a complementary question: \emph{what representation of the 3D physical scene should such policies reason over?}

\begin{figure}[t]
\centering
\includegraphics[width=\textwidth]{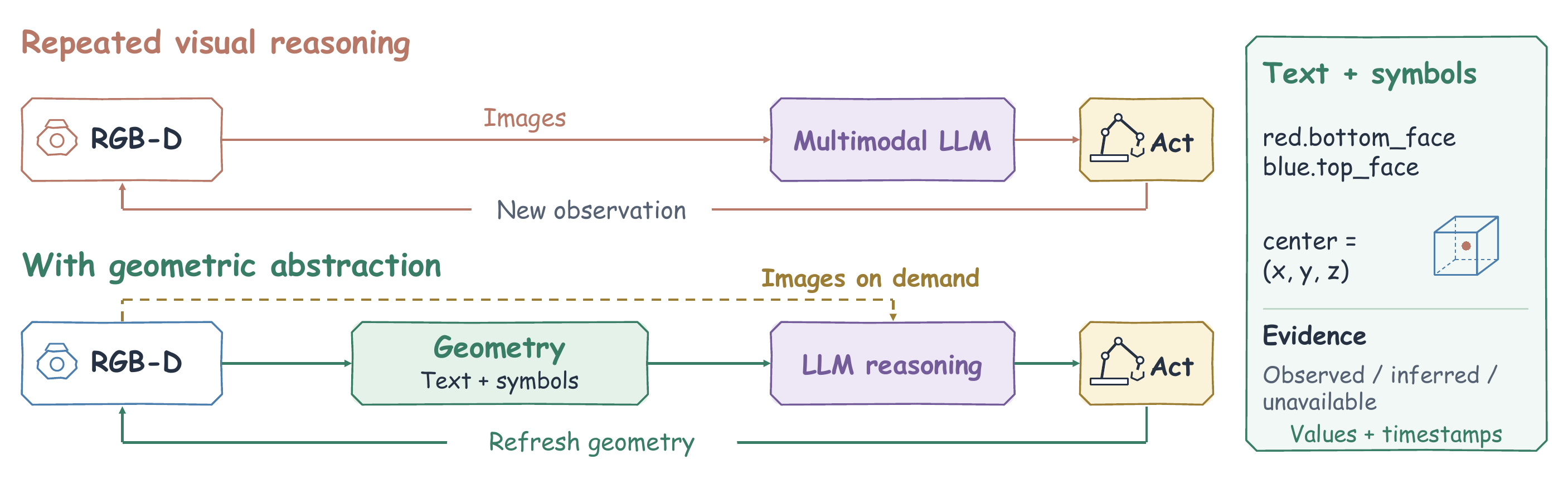}
\caption{\textbf{Geometric abstraction between observation, reasoning, and
action.} Visual observations are grounded locally into metric 3D geometry and exposed
to the language model through symbolic references and compact textual checkpoints.
Policies reason in text and symbols over stable geometric roles whose metric values are rebound to the current
scene during execution.}
\label{fig:overview}
\end{figure}

To address this question, we propose shifting routine spatial reasoning from images to \emph{text and symbols} through an explicit \emph{3D geometric abstraction} (Fig.~\ref{fig:overview}). Rather than reducing objects to single 3D centroids or vague natural-language descriptions, our abstraction models scene geometry as a structured set of typed 3D object parts---including centers, vertices, edges, and oriented surfaces---annotated with timestamps and evidence status (\emph{observed}, \emph{inferred} from shape priors, or \emph{unavailable} under occlusion). Local perception grounds this 3D state from calibrated RGB-D and exposes it to the language model in two forms. First, \emph{symbolic references} (e.g., \texttt{red.bottom\_face}, \texttt{blue.top\_face}) and geometric operators (such as surface alignment and support checks) let policies compose actions independently of specific object coordinates. Second, \emph{compact textual checkpoints} report updated 3D coordinates, metric goal residuals, and evidence status after each action. Separating reusable symbolic structure from scene-dependent metric grounding lets the language model reason directly in text and symbols, invoking visual reasoning on demand when richer scene context is needed.

We study this textual and symbolic 3D interface in two regimes. In \emph{policy synthesis}, the language model revises an executable policy between rollouts using execution feedback, while geometric grounding and control run locally during rollouts. In \emph{online execution}, the agent repeatedly reasons and selects skills within a single continuing episode using the updated textual checkpoint after each action, without resetting the scene after failures. Both settings use matched sensors, actuators, and initial skills. Across both regimes, our experiments show that reasoning over grounded text and symbols accelerates policy acquisition, substantially improves multi-step online construction completion, and reduces repeated multimodal queries and interaction latency. We use these two settings to examine when explicit textual and symbolic 3D state is sufficient for manipulation and when direct visual reasoning remains necessary.

Our contributions are threefold: (i) we formulate a 3D geometric abstraction that
exposes task-relevant scene geometry to language models as structured text and symbols---combining typed, evidence-aware part references with reusable symbolic structure and local metric grounding;
(ii) we instantiate this abstraction as a reasoning-to-action interface with composable geometric
operators, local RGB-D grounding, and compact textual checkpoints; and (iii) we
empirically evaluate the interface across policy synthesis and continuous online execution,
characterizing its impact on manipulation completion, model interaction cost, and the
remaining role of direct visual input.

\section{Related Work}

\paragraph{Geometric representations and spatial reasoning for manipulation.}
Prior work bridges high-level reasoning and low-level robot control through structured intermediate representations. VoxPoser~\citep{huang2023voxposer} synthesizes composable 3D value maps for motion planning, ReKep~\citep{huang2024rekep} formulates manipulation as relational constraints over sparse 3D keypoints, and KALM~\citep{fang2024kalm}, RT-Affordance~\citep{nasiriany2025rtaffordance}, and ATK~\citep{zhang2025atk} extract task-driven keypoints or affordances to guide policy learning. Meanwhile, recent VLM evaluations show that recovering metric and structural 3D relations directly from images remains error-prone and token-intensive, whereas making spatial relations explicit in text improves reasoning reliability~\citep{wang2024picture,pothiraj2025capture,song2025siri,yang2025thinking,chen2024image}. Rather than optimizing a single-step spatial objective or relying on coarse centroids, we maintain a persistent, evidence-aware \emph{3D scene state} of typed object parts (centers, vertices, edges, and oriented faces) that is continuously re-grounded from RGB-D and exposed to the language model as composable symbols and compact text.

\paragraph{Programmatic, agentic, and closed-loop robot policies.}
Code as Policies~\citep{liang2022code} and Inner Monologue~\citep{huang2022inner} pioneered composing perception-control APIs in code and incorporating closed-loop language feedback. Recent agentic frameworks extend this paradigm through coding-agent benchmarks and VLM module orchestration~\citep{fu2026capx,shi2025maestro}, modular policy graphs and evolving skill libraries~\citep{chen2026gap,fu2026segos}, execution-driven repair in simulation and the real world~\citep{lu2026aspire,xiao2026enpire}, and hierarchical VLA steering and failure correction~\citep{zhang2026harnessvla,hu2026orchestrating,kwon2026correctvla}. While these works focus on how policies are synthesized, orchestrated, or repaired, we focus on the \emph{spatial state interface} that closes the loop: re-grounding symbolic references after each action and returning metric goal residuals and evidence status in text so subsequent decisions reason over the observed post-action scene rather than assumed action outcomes.

\section{Geometric Abstraction as a Reasoning-to-Action Interface}
\label{sec:method}

\begin{figure}[t]
\centering
\includegraphics[width=\textwidth]{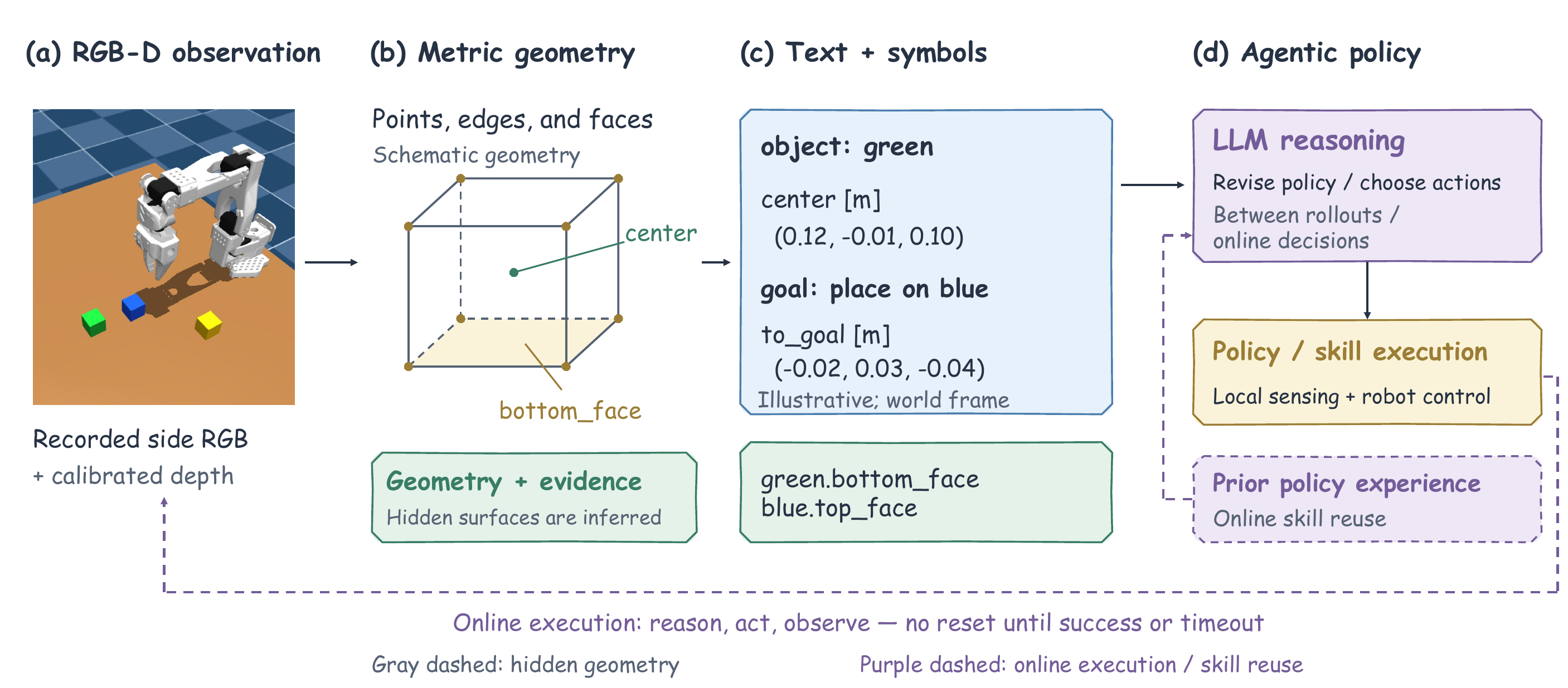}
\caption{\textbf{Geometric abstraction between perception and reasoning.}
RGB-D observations are grounded into object geometry and exposed to the LLM
through typed symbolic references and compact textual checkpoints.
In policy synthesis, the LLM revises executable policies between rollouts,
while perception and control run locally. In online execution, the same
interface supplies updated geometry for successive decisions within a
continuing episode.}
\label{fig:pipeline}
\end{figure}

Our interface represents scene geometry through typed symbolic references that are
grounded in current RGB-D observations at execution time. Updated geometry and
goal residuals are returned as compact textual checkpoints for subsequent
reasoning (Fig.~\ref{fig:pipeline}).

\subsection{Grounding geometry from visual observations}
\label{sec:state}

At time $t$, calibrated RGB-D observations from the side and wrist cameras,
together with robot proprioception $q_t$, define the current geometric state:
\begin{equation}
o_t=
\big(
\{I_t^{c},D_t^{c},K^{c},T_{WC}^{c}\}_{c\in\{\mathrm{side},\mathrm{wrist}\}},
q_t
\big),
\qquad
\mathcal Z_t=\Phi(o_t;M),
\label{eq:obs}
\end{equation}
where $M$ stores object shape information acquired during initialization.
Depth observations are back-projected into the calibrated world frame and combined
with $M$ to estimate each object's current center, orientation, dimensions, and,
when supported by the observation, its vertices, edges, and faces. Each geometric
element additionally records whether it is directly observed, inferred from the
initialized shape model, or unavailable. A core invariant is that \emph{shape may persist, but pose must be refreshed}: after an
object moves, its pose must be re-estimated from a current observation rather than
copied from the previous state or inferred from commanded motions. If the
required geometry cannot be recovered, the corresponding reference is marked
unavailable.

\subsection{Symbolic references and geometric operators}
\label{sec:references}

Policies written by the language model access the geometric state through typed references. A reference $r=(i,\rho)$ identifies object $i$ and a geometric role $\rho$ (such as a center, vertex, segment, or face) and is evaluated against the current state:
\begin{equation}
\llbracket i.\rho\rrbracket_{\mathcal Z_t}
=
\begin{cases}
(v,\tau,\eta,t),
& \text{if the reference is currently supported},\\
\bot,
& \text{otherwise},
\end{cases}
\label{eq:reference}
\end{equation}
where $v$ is the metric geometric value, $\tau$ its type, $\eta$ its evidence status, and $t$ its observation timestamp.

This turns scene geometry into a compositional vocabulary (Table~\ref{tab:operators}). Instead of reasoning directly from images, a policy can refer to \texttt{red.bottom\_face} and \texttt{blue.top\_face}, align the two faces, and request the corresponding placement transform. The same program generalizes across object configurations because symbolic references are dynamically rebound to the current scene geometry.

\begin{table}[t]
\centering
\caption{Operators on the geometric state.}
\label{tab:operators}
\small
\begin{tabular}{@{}lll@{}}
\toprule
Operation & Inputs & Output\\
\midrule
\texttt{reference}
& object and role
& geometric entity with evidence\\
\texttt{relation}
& predicate and objects
& residuals and \texttt{true}/\texttt{false}/\texttt{unknown}\\
\texttt{align}
& source and target geometry
& desired rigid transform\\
\texttt{checkpoint}
& active object and goal
& compact textual state\\
\bottomrule
\end{tabular}
\end{table}

For placement, the policy queries the mating face centers and outward normals $(p_a,n_a)=\mathrm{bottom}(a)$ and $(p_b,n_b)=\mathrm{top}(b)$ and computes a rigid transform $(R_\Delta, t_\Delta)$ satisfying
\begin{equation}
R_\Delta n_a=-n_b,
\qquad
t_\Delta=p_b-R_\Delta p_a.
\label{eq:align}
\end{equation}
Since a point and a normal leave rotation about $n_b$ unconstrained, \texttt{align} returns the minimum-angle rotation and reports the remaining twist. This provides a reusable geometric primitive rather than a hardcoded coordinate target. Spatial relations are evaluated similarly: a support query checks relative height, planar overlap, and face orientation, returning both the relation status and its metric residuals. The status is \texttt{unknown} when a required axis or surface is unobserved, and a \texttt{true} result is treated as a geometric candidate rather than a certificate of physical contact or stability.

\subsection{Textual checkpoints for reasoning}
\label{sec:checkpoint}

Symbolic references support action composition, while textual checkpoints expose the current state in a compact form for reasoning. For active object $i$ and target position $g_i$, we write
\begin{equation}
x_t=\Psi(\mathcal Z_t,i,g_i),
\qquad
e_t=g_i-p_i(t),
\label{eq:checkpoint}
\end{equation}
where $x_t$ summarizes the relevant current geometry, its evidence status, and the residual $e_t$ between $g_i$ and the current position $p_i(t)$ of the selected reference. For example, after placing a block, the checkpoint reports whether the target face is aligned, the remaining translational displacement, and whether the current estimate is visually supported. If the object becomes occluded, the goal is retained but the residual is explicitly marked unavailable rather than computed from a stale pose estimate.

This representation allows successive reasoning steps to reuse grounded geometry without repeatedly sending raw images to the language model, while keeping visual observations available on demand when new scene context is needed or when the geometric abstraction does not explain a failure.

\subsection{Policy synthesis and online execution}
\label{sec:usage}

We evaluate the same interface in two operational regimes. In \emph{policy synthesis} (single-policy acquisition), the
language model revises an executable program between batches of rollouts; within
each rollout, geometry is grounded locally and the program executes without
further language-model calls.

In \emph{online execution}, the language model instead participates directly in the
continuing interaction loop:
\begin{equation}
u_t=\mathcal A_{\mathrm{LLM}}(g,x_t,\mathcal H_t,\mathcal E),
\end{equation}
where $x_t$ is the current textual checkpoint, $\mathcal H_t$ the interaction history, and
$\mathcal E$ available prior skill experience. After each action, new
observations re-ground the relevant references, ensuring that recovery decisions start from the
observed post-action scene rather than from the intended action outcome. Images remain
available on demand when the geometric state is insufficient.

\section{Experiments}
We evaluate the geometric abstraction in simulation with an SO-100 robot arm
equipped with calibrated side and wrist RGB-D cameras. Across all experiments,
the agent receives the same task specification, sensor access, actuator interface, and
initial skills; no simulator object poses or private evaluator state are exposed.
We compare a \emph{Pixel Reasoning} agent that reasons directly from visual observations with a
\emph{Geometric Abstraction} agent that additionally receives our grounded geometric references and
textual checkpoints. We study the interface in two regimes: policy synthesis
across repeated rollouts and continuous online execution within a single,
non-resetting episode.

\subsection{Single-policy acquisition}
\label{sec:single-policy-experiments}

\begin{figure*}[t]
  \centering
  \includegraphics[width=\textwidth]{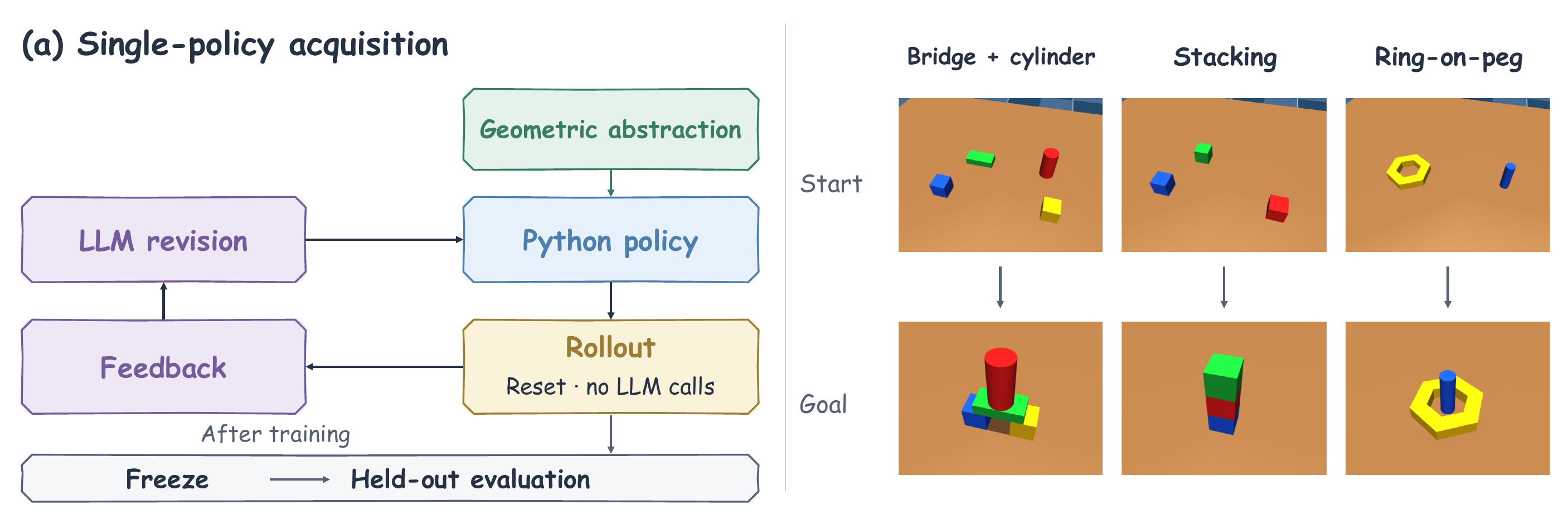}
  \caption{\textbf{Single-policy acquisition.} Policies are revised between reset
  rollouts and evaluated after training freezes. Geometric Abstraction additionally uses
  supplied geometric references. Right: illustrative start and goal configurations,
  constructed in simulation rather than recorded policy trajectories.}
  \label{fig:single-policy-task-illustrations}
\end{figure*}

\begin{figure*}[t]
  \centering
  \includegraphics[width=\textwidth]{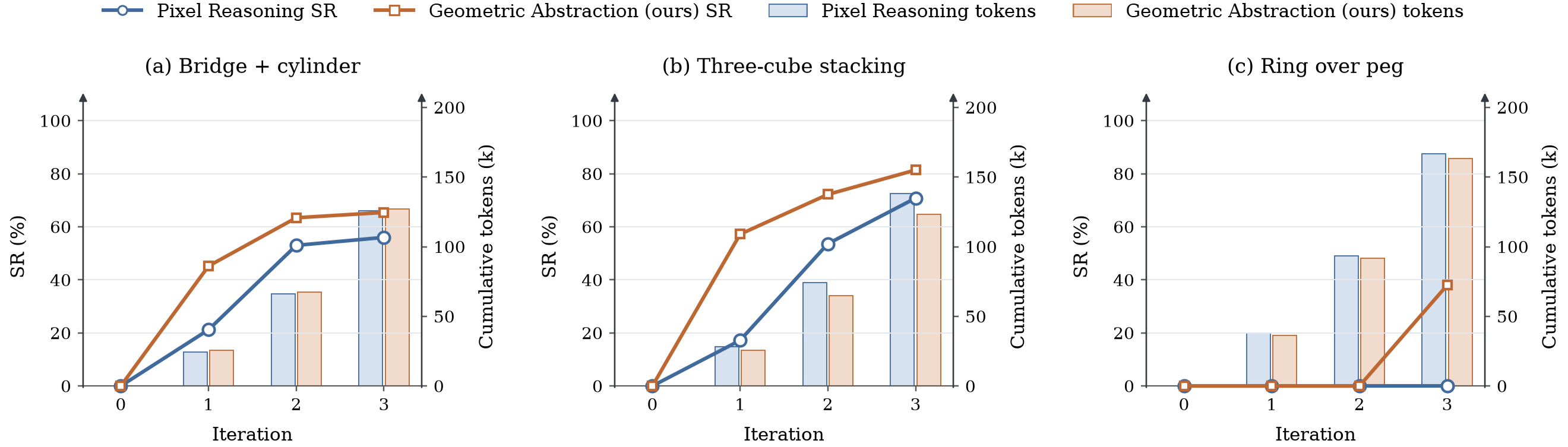}
  \caption{\textbf{Test success and acquisition cost.} Lines show mean held-out test SR of
  training-retained policies; bars show cumulative input-plus-output tokens
  through each iteration, including rejected proposals. Both average over five
  chains. Ring-on-peg Pixel Reasoning excludes unknown usage from two interrupted requests.}
  \label{fig:single-policy-curves}
\end{figure*}

In the policy synthesis regime, we test whether geometric abstraction helps an LLM acquire effective single-program policies within a small revision budget. The simulated SO-100 performs three manipulation tasks with randomized object poses and colors (Figure~\ref{fig:single-policy-task-illustrations}): \emph{Bridge+cylinder} (placing two base blocks, spanning them with a beam, and placing an upright cylinder on top), \emph{Three-cube stacking} (assembling a three-cube tower in a prescribed color order), and \emph{Ring-on-peg} (grasping a hexagonal ring and threading it onto a vertical peg).

\paragraph{Comparison and procedure.}
Pixel Reasoning and Geometric Abstraction share the task specification, calibrated RGB-D access, and action interface, both using \texttt{gpt-6-astra} with high reasoning effort. Pixel Reasoning implements custom perception routines directly in its generated program, whereas Geometric Abstraction additionally queries sensor-grounded 3D part references, geometric operators, and textual checkpoints. Neither condition has access to simulator ground-truth poses or private evaluator state.

For each task, we run five paired training chains over five shared training layouts with separate persistent LLM contexts. Starting from an observation-only scaffold (R0), the LLM generates three policy revisions (R1--R3). Each revision executes across all five training layouts with a scene reset between rollouts and no LLM calls during execution; rollout outcomes and observations inform the subsequent revision. A candidate policy is retained only if it succeeds on every previously solved training layout plus at least one additional layout; otherwise, the incumbent policy is kept. After training freezes, each retained version is evaluated on 100 shared held-out layouts per task without further revision. Physical success requires achieving the target arrangement, releasing the object, and maintaining stability (Figure~\ref{fig:single-policy-curves}; Table~\ref{tab:single-policy-results} in Appendix~\ref{app:single-policy-details}).

\paragraph{Findings.}
Geometric abstraction substantially accelerates policy acquisition on both construction tasks. The largest gap appears at the first revision (R1), where Geometric Abstraction reaches $45.2\%$ held-out test SR on Bridge+cylinder (vs.\ $21.2\%$ for Pixel Reasoning) and $57.2\%$ on Three-cube stacking (vs.\ $17.2\%$). By R2, Geometric Abstraction's mean test SR ($63.4\%$ and $72.2\%$) already surpasses Pixel Reasoning's final R3 performance ($56.0\%$ and $70.8\%$) on both tasks, while expending roughly half the cumulative tokens ($67.3\text{k}$ vs.\ $125.7\text{k}$ on Bridge+cylinder; $64.7\text{k}$ vs.\ $138.1\text{k}$ on Stacking). Although the gap narrows at R3 as Pixel Reasoning refines its perception code, Geometric Abstraction retains higher final success ($65.4\%$ and $81.4\%$) and lower chain-to-chain variance, with comparable total tokens on Bridge+cylinder and lower token usage on Stacking. Ring-on-peg follows a different pattern: both conditions remain at $0.0\%$ retained test SR through R2, before Geometric Abstraction reaches $38.0\%$ at R3 while Pixel Reasoning remains at $0.0\%$. Here, geometric abstraction enables successful policy acquisition within the three-revision budget where image-only synthesis fails to converge, as fine ring-to-peg alignment benefits directly from grounded part references. Note that these gains reflect the combined effect of policy synthesis and local geometric grounding during execution, rather than isolating LLM inference speed or text formatting alone.

\subsection{Continuous online reasoning and execution}
\label{sec:online-experiments}

\begin{figure*}[t]
  \centering
  \includegraphics[width=\textwidth]{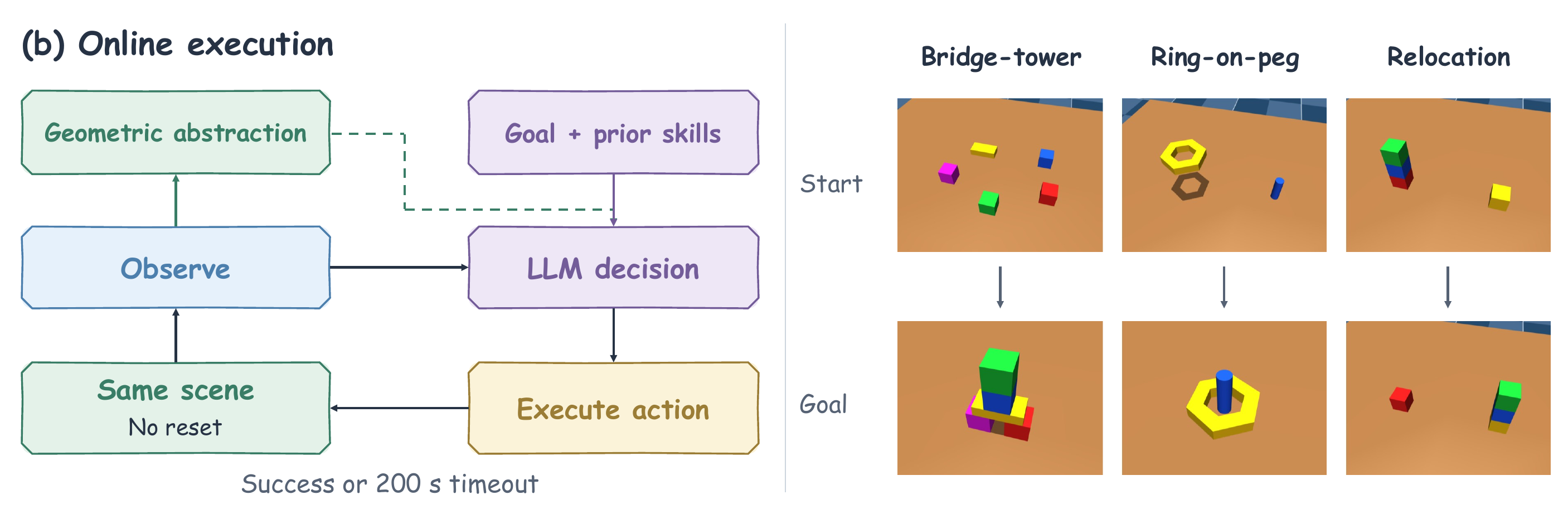}
  \caption{\textbf{Online execution and tasks.} The agent acts in a single
  continuing scene until completion or the 200\,s timeout; Geometric Abstraction receives grounded
  symbolic references and textual checkpoints. Right: illustrative start and goal states.}
  \label{fig:online-task-illustrations}
\end{figure*}

\begin{figure*}[t]
  \centering
  \includegraphics[width=\textwidth]{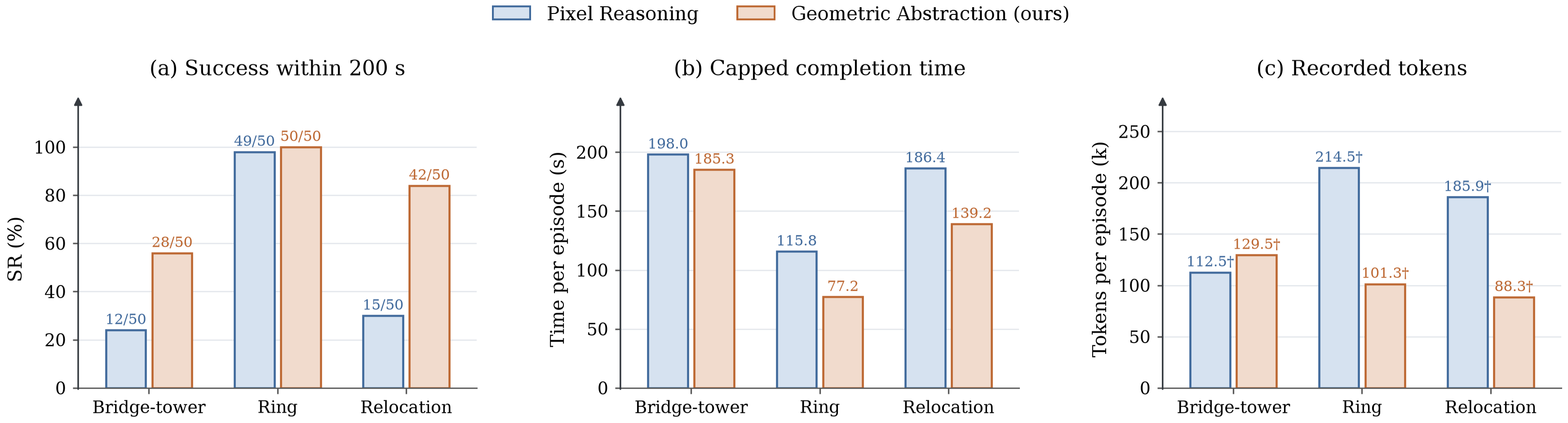}
  \caption{\textbf{Online outcomes and cost.} Completion rate within 200\,s (left),
  mean capped completion time (middle; timeouts assigned 200\,s), and mean recorded tokens
  (right) over 50 paired layouts per task ($\dagger$: excludes unknown interrupted tokens).}
  \label{fig:online-outcomes}
\end{figure*}

We next evaluate whether geometric abstraction helps an LLM (\texttt{gpt-6-astra}, medium reasoning effort) orchestrate prior skills and recover from failures online under a real-time wall-clock budget (Figure~\ref{fig:online-task-illustrations}). Failed actions are not reset, so model waiting time shares the same 200-second episode budget as physical execution and recovery.

\paragraph{Tasks and protocol.}
We evaluate 50 paired layouts per task (300 episodes total) across three tasks: \emph{Bridge-tower} (stacking two cubes atop a constructed bridge), \emph{Ring-on-peg} (aligning and releasing a pre-lifted ring over a peg), and \emph{Tower relocation} (transferring the upper two cubes of a column onto a new foundation while preserving order and foundation poses). Both conditions share identical local skills (bridge/stacking primitives, a shared sensor-driven ring grasp-and-lift prefix, and placement routines) and can request camera images and depth samples on demand; Geometric Abstraction additionally receives compact textual checkpoints and symbolic references after each action. Each episode runs continuously for up to 200\,s with physics advancing during model waits (see Appendix~\ref{app:online-details}--\ref{app:geometric-abstraction} for full protocol and audited repeat details).

\paragraph{Findings.}
Geometric Abstraction substantially increases completion rates on multi-step construction, solving Bridge-tower on 28/50 layouts (56\% vs.\ 12/50, 24\%) and Tower relocation on 42/50 layouts (84\% vs.\ 15/50, 30\%) with lower mean capped completion times (Figure~\ref{fig:online-outcomes}; Table~\ref{tab:online-results} in Appendix~\ref{app:online-details}). On Ring-on-peg, where both conditions reliably succeed (50/50 vs.\ 49/50), Geometric Abstraction finishes 33.3\% faster (77.2\,s vs.\ 115.8\,s), completing earlier on 45 of 50 pairs.

\begin{figure*}[t]
  \centering
  \includegraphics[width=\textwidth]{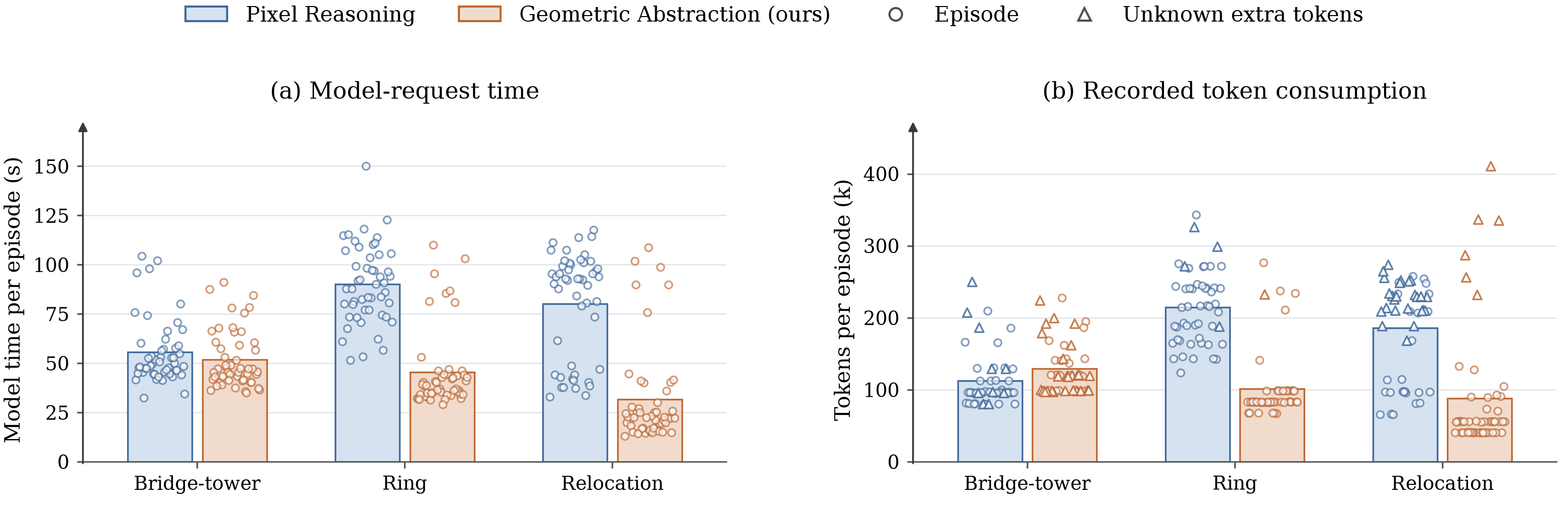}
  \caption{\textbf{Model interaction cost.} Mean (bars) and per-episode (markers) request wall-clock time (left) and recorded tokens (right) over 50 paired episodes per task ($\triangle$: excludes unknown interrupted tokens).}
  \label{fig:online-model-resources}
\end{figure*}

Simultaneously, Geometric Abstraction reduces repeated multimodal querying (Figure~\ref{fig:online-model-resources}). On Ring-on-peg, model-request time drops by 49.4\% (45.6\,s vs.\ 90.0\,s), recorded tokens by 52.8\% (101.3k vs.\ 214.5k), completed requests from 11.88 to 6.90, and attached images from 10.94 to 2.72. Tower relocation shows a 60.5\% reduction in model time and 52.5\% in tokens while physical execution time remains similar (104.5\,s vs.\ 103.2\,s). On Bridge-tower, Geometric Abstraction more than doubles completion with 6.7\% less model time and 15.1\% more tokens, showing that the interface improves performance both by cutting visual query overhead and by enabling effective corrective actions within the time budget. Across all tasks, local RGB-D sensing refreshes the 3D state while drastically reducing image attachments to the LLM (Table~\ref{tab:online-results}; Figure~\ref{fig:online-success-time}).

\section{Limitations and Discussion}
\label{sec:discussion}

Our evaluation focuses on simulated table-top manipulation with regular
object geometries. Real-world deployment remains an essential next step to assess robustness
under realistic sensor noise, heavy occlusion, and camera calibration drift. We deliberately used
lightweight CPU-based geometric estimators to keep local grounding fast and self-contained, leaving learned 3D perception modules
and articulated or deformable objects for future work.

In continuous online execution, geometric abstraction reduced decision rounds and on-demand image
attachments across tasks, indicating that its primary benefit lies in providing direct, actionable spatial state
throughout the reasoning--action loop rather than merely compressing prompt length. More concise checkpoint schemas and learned
symbolic abstractions could further reduce interaction overhead. For novel object categories,
VLM-assisted shape initialization could broaden coverage, provided its one-time grounding cost is
accounted for alongside policy execution and amortized across subsequent interactions.

Finally, decoupling local visual grounding from high-level policy reasoning opens the possibility of driving manipulation with fast text-only language models.
By exposing verified, sensor-grounded 3D state in text, the interface allows a language model to reason over physical geometry without processing dense visual tokens at every step.
Evaluating diverse models on standardized geometric checkpoints could clarify
how effectively language models perform spatial reasoning when metric relations are made explicit,
and help delineate which manipulation decisions are well served by structured geometric state versus direct visual inspection.

\section{Conclusion}
\label{sec:conclusion}

We investigated whether agentic robot policies need to reinterpret visual observations at every
decision step. Across both offline policy synthesis and continuous online execution, our results
demonstrate that maintaining an explicit geometric abstraction improves manipulation completion while
substantially reducing repeated multimodal queries. By pairing reusable symbolic references with
metric values refreshed from local RGB-D observations, the interface allows visual sensing to remain
continuously active in the control loop without requiring the language model to process images at every
step---positioning structured geometric state as a practical substrate for agentic reasoning
and action, with direct visual reasoning invoked on demand.

\clearpage
\bibliography{references}
\clearpage
\appendix
\input{sections/appendix}

\end{document}

%% file: sections/appendix.tex
\section{Additional single-policy results}
\label{app:single-policy-details}

\begin{table*}[t]
  \centering
  \small
  \setlength{\tabcolsep}{4pt}
  \resizebox{\textwidth}{!}{%
  \begin{tabular}{llrrrrrr}
    \toprule
    & & \multicolumn{3}{c}{Retained test SR (\%)}
      & \multicolumn{3}{c}{Cumulative tokens (k)} \\
    \cmidrule(lr){3-5}\cmidrule(lr){6-8}
    Task & Condition & R1 & R2 & R3 & R1 & R1--R2 & R1--R3 \\
    \midrule
    Bridge + cylinder & Pixel Reasoning & $21.2 \pm 22.0$ & $53.0 \pm 11.7$ & $56.0 \pm 9.1$ & 24.6 & 66.2 & 125.7 \\
     & Geometric Abstraction (ours) & $\boldsymbol{45.2 \pm 25.4}$ & $\boldsymbol{63.4 \pm 5.9}$ & $\boldsymbol{65.4 \pm 3.5}$ & 25.8 & 67.3 & 127.0 \\
    \midrule
    Three-cube stacking & Pixel Reasoning & $17.2 \pm 29.4$ & $53.6 \pm 26.2$ & $70.8 \pm 26.5$ & 28.0 & 74.1 & 138.1 \\
     & Geometric Abstraction (ours) & $\boldsymbol{57.2 \pm 32.6}$ & $\boldsymbol{72.2 \pm 7.2}$ & $\boldsymbol{81.4 \pm 12.3}$ & \textbf{25.7} & \textbf{64.7} & \textbf{123.4} \\
    \midrule
    Ring-on-peg & Pixel Reasoning & $0.0 \pm 0.0$ & $0.0 \pm 0.0$ & $0.0 \pm 0.0$ & 37.8 & 93.5 & 166.8 \\
     & Geometric Abstraction (ours) & $0.0 \pm 0.0$ & $0.0 \pm 0.0$ & $\boldsymbol{38.0 \pm 34.9}$ & \textbf{36.3} & \textbf{91.5} & \textbf{163.1} \\
    \bottomrule
  \end{tabular}}
  \caption{\textbf{Single-policy acquisition outcomes and token usage.} Held-out test SR (mean $\pm$ sample SD) across five training chains on 100
  shared held-out layouts per task, using training-only policy retention. Tokens report
  cumulative completed-request input plus output per chain (in thousands), including
  rejected proposals and counting cached input once. Ring-on-peg Pixel Reasoning excludes unknown
  token usage from two interrupted requests. Bold highlights where Geometric Abstraction (ours) improves over Pixel Reasoning.}
  \label{tab:single-policy-results}
\end{table*}

\paragraph{Evaluation notes.}
All experiments use MuJoCo 3.11.0~\citep{todorov2012mujoco} with a simulated SO-100 robot arm. Bridge+cylinder and Three-cube stacking require achieving the prescribed support structure and maintaining a five-second
stable hold after release. Ring-on-peg requires the peg to pass through the ring opening, the ring center to lie below
the peg top, full release from robot contact, and a two-second stable hold. Syntactically or runtime-invalid
programs count as failures. Bridge+cylinder and Stacking contact checks can miss unnamed robot collision
geometries, whereas Ring-on-peg uses full body-ancestry contact checks.

\section{Additional online results}
\label{app:online-details}

\begin{figure*}[t]
  \centering
  \includegraphics[width=\textwidth]{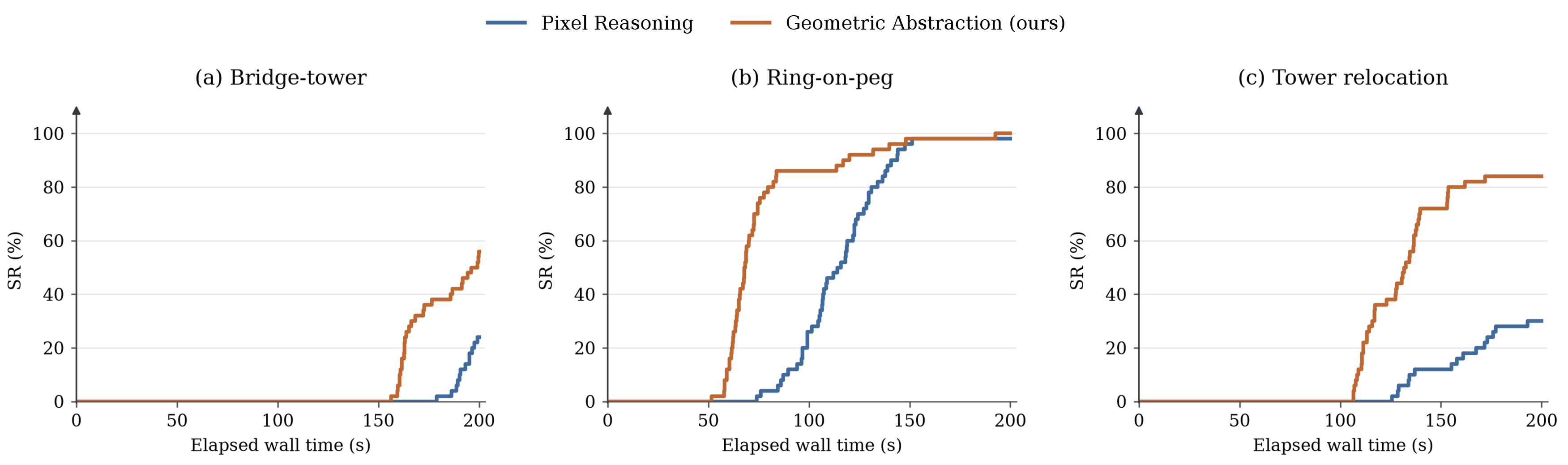}
  \caption{\textbf{Cumulative success over elapsed wall-clock time.} Each curve summarizes the 50
  paired 200-second episodes for Pixel Reasoning and Geometric Abstraction (ours), stepping at the first verified completion time. Timeouts
  do not convert to successes at the 200\,s deadline. Intermediate values are read directly from the
  200-second episodes rather than separate evaluations under shorter budgets.}
  \label{fig:online-success-time}
\end{figure*}

Table~\ref{tab:online-results} and Figure~\ref{fig:online-success-time} expand the online evaluation results
presented in Section~\ref{sec:online-experiments}.

\begin{table*}[t]
  \centering
  \small
  \setlength{\tabcolsep}{4pt}
  \resizebox{\textwidth}{!}{%
  \begin{tabular}{llrrrrrr}
    \toprule
    Task & Condition & Success & Time (s) & Model (s) & Tokens (k) & Requests & Images \\
    \midrule
    Bridge-tower & Pixel Reasoning & 12/50 & 198.0 & 55.5 & 112.5 & 7.88 & 7.50 \\
     & Geometric Abstraction (ours) & \textbf{28/50} & \textbf{185.3} & \textbf{51.8} & 129.5 & \textbf{7.32} & \textbf{4.28} \\
    \midrule
    Ring-on-peg & Pixel Reasoning & 49/50 & 115.8 & 90.0 & 214.5 & 11.88 & 10.94 \\
     & Geometric Abstraction (ours) & \textbf{50/50} & \textbf{77.2} & \textbf{45.6} & \textbf{101.3} & \textbf{6.90} & \textbf{2.72} \\
    \midrule
    Tower relocation & Pixel Reasoning & 15/50 & 186.4 & 80.3 & 185.9 & 11.58 & 11.34 \\
     & Geometric Abstraction (ours) & \textbf{42/50} & \textbf{139.2} & \textbf{31.7} & \textbf{88.3} & \textbf{5.36} & \textbf{3.00} \\
    \bottomrule
  \end{tabular}}
  \caption{\textbf{Online outcomes and interaction cost.} Each task evaluates 50 paired layouts.
  Success reports completed episodes out of 50; all other columns report per-episode
  means across all 50 episodes, including timeouts. Time is capped completion time, assigning
  200 seconds to timeouts. Model is cumulative request wall-clock time, including
  interrupted waits. Tokens are recorded input plus output tokens (in thousands),
  counting cached input once and excluding unknown interrupted usage.
  Requests counts completed model calls; Images counts newly attached camera
  images across all request attempts (excluding local sensing frames not sent to the LLM).
  Both conditions use audited repeats for nine runtime-flagged pairs. Bold highlights where Geometric Abstraction (ours) improves over Pixel Reasoning.}
  \label{tab:online-results}
\end{table*}

\paragraph{Evaluation notes.}
Bridge-tower and Tower relocation require a five-second stable hold; Ring-on-peg requires two
seconds. Tower relocation must preserve or restore both foundation cubes within 5\,mm of
their initial positions. Ring-on-peg executes the same sensor-driven grasp-and-lift prefix in
both conditions and hands control to the online agent prior to alignment and release; completion times and
all interaction costs include this shared prefix. Each layout is attempted once per condition
(with audited repeats replacing nine pairs affected by runtime timing or GPU faults). Because
paired episodes can differ in outcome, completion-time means include all 50 episodes (with timeouts capped at 200\,s) rather than conditioning on successful episodes only. Model-request time reflects end-to-end API waiting time
rather than isolated model compute.

\section{Geometric abstraction implementation}
\label{app:geometric-abstraction}

The geometric abstraction converts calibrated RGB-D observations into metric entities, symbolic
references, and compact textual checkpoints. It runs locally on the CPU using fixed geometric estimators
throughout both policy synthesis and online execution.

\paragraph{From pixels to metric geometry.}
Color segmentation masks associate visible pixels with scene objects. For a valid pixel
$\tilde{\mathbf u}=(u,v,1)^\top$, its 3D world position is recovered via back-projection:
\begin{equation}
  \mathbf p_W=R_{WC}\bigl(D(u,v)K^{-1}\tilde{\mathbf u}\bigr)+\mathbf t_{WC},
\end{equation}
where $D(u,v)$ is optical-axis depth, $K$ is the camera intrinsic matrix, and
$(R_{WC},\mathbf t_{WC})$ transforms camera coordinates to the world frame. Background depth points
estimate the tabletop plane, while known robot link geometry masks out the manipulator.
Cuboid and cylinder estimators fit object pose and visible support surfaces using dimensions
initialized from RGB-D under declared shape and table-support priors.
Simulator ground-truth object poses and dimensions are never accessed.

\paragraph{References and current evidence.}
Symbolic references such as \texttt{red.top\_face} resolve to typed geometric primitives: points, axes, or
oriented faces with centers and outward normals. Each record stores world-frame coordinates, physical units,
observation timestamps, and evidence status (\texttt{observed}, \texttt{inferred}, or \texttt{unavailable}). Initialized shape dimensions
persist across an episode, whereas a movable object's pose must be refreshed from current visual evidence after movement.
For partially occluded rings, initialized boundary templates are registered to currently
visible contour edges; underconstrained or ambiguous fits return \texttt{unavailable}. Unobserved object
faces retain model-derived labels marked as inferred, while the stationary peg may retain an earlier
verified measurement with its original timestamp. Geometric support predicates report metric
residuals and missing evidence rather than certifying physical contact or dynamic stability.

\paragraph{From geometry to agent input.}
The Geometric Abstraction agent receives task-relevant object geometry, symbolic references, evidence metadata, and
requested goal residuals formatted as structured text (with dense boundary point arrays omitted from online
Ring-on-peg checkpoints for compactness). Multiple geometric queries can reuse a single RGB-D capture, and local sensing
can refresh the geometric state after an action without attaching new images to the LLM prompt. Initial and
on-demand camera images remain accessible whenever visual inspection is requested. In policy synthesis, generated programs invoke geometric operators
locally during rollouts; in online execution, the agent reads updated textual checkpoints to select its next skill or recovery action.